\documentclass{article}
\usepackage{spconf,amsmath,graphicx,hyperref}

\usepackage{subcaption}
\usepackage{booktabs}
\usepackage{multirow}
\usepackage{tikz}
\usepackage{pgfplots}
\pgfplotsset{compat=1.18}

\newcommand\copyrighttext{%
  \footnotesize \textcopyright 2026 IEEE. Personal use of this material is permitted.
  Permission from IEEE must be obtained for all other uses, in any current or future
  media, including reprinting/republishing this material for advertising or promotional
  purposes, creating new collective works, for resale or redistribution to servers or
  lists, or reuse of any copyrighted component of this work in other works. }
    \newcommand\mycopyrightnotice{%
\begin{tikzpicture}[remember picture,overlay]
\node[anchor=south,yshift=10pt] at (current page.south) {\fbox{\parbox{\dimexpr\textwidth-\fboxsep-\fboxrule\relax}{\copyrighttext}}};
\end{tikzpicture}%
}

\usepackage{xcolor}

\definecolor{dc1}{RGB}{230, 25, 75} 
\definecolor{dc2}{RGB}{60, 180, 75} 
\definecolor{dc3}{RGB}{255, 225, 25} 
\definecolor{dc4}{RGB}{0, 130, 200} 
\definecolor{dc5}{RGB}{245, 130, 48} 
\definecolor{dc6}{RGB}{145, 30, 180} 
\definecolor{dc7}{RGB}{70, 240, 240} 
\definecolor{dc8}{RGB}{240, 50, 230} 
\definecolor{dc9}{RGB}{210, 245, 60} 
\definecolor{dc10}{RGB}{250, 190, 212} 
\definecolor{dc11}{RGB}{0, 128, 128} 
\definecolor{dc12}{RGB}{220, 190, 255} 
\definecolor{dc13}{RGB}{170, 110, 40} 
\definecolor{dc14}{RGB}{255, 250, 200} 
\definecolor{dc15}{RGB}{128, 0, 0} 
\definecolor{dc16}{RGB}{170, 255, 195} 
\definecolor{dc17}{RGB}{128, 128, 0} 
\definecolor{dc18}{RGB}{255, 215, 180} 
\definecolor{dc19}{RGB}{0, 0, 128} 
\definecolor{dc20}{RGB}{128, 128, 128} 
\definecolor{dc21}{RGB}{255, 255, 255} 
\definecolor{dc22}{RGB}{0, 0, 0} 

\title{Chain-of-caption: training-free improvement of multimodal large language model on referring expression comprehension}
\name{Yik Lung Pang, Changjae Oh \thanks{This work was supported by the Korea Institute for Advancement of Technology (KIAT) grant (P0028485, GITCC).}}
\address{Queen Mary University of London}

\begin{document}
\ninept
\maketitle
\begin{abstract}
Given a textual description, the task of referring expression comprehension (REC) involves the localisation of the referred object in an image. Multimodal large language models (MLLMs) have achieved high accuracy on REC benchmarks through scaling up the model size and training data. Moreover, the performance of MLLMs can be further improved using techniques such as Chain-of-Thought and tool use, which provides additional visual or textual context to the model. In this paper, we analyse the effect of various techniques for providing additional visual and textual context via tool use to the MLLM and its effect on the REC task. Furthermore, we propose a training-free framework named Chain-of-Caption to improve the REC performance of MLLMs. We perform experiments on RefCOCO/RefCOCOg/RefCOCO+ and Ref-L4 datasets and show that individual textual or visual context can improve the REC performance without any fine-tuning. By combining multiple contexts, our training-free framework shows between 5\% to 30\% performance gain over the baseline model on accuracy at various Intersection over Union (IoU) thresholds.
\end{abstract}
\mycopyrightnotice
\section{Introduction}
\label{sec:intro}

The task of referring expression comprehension (REC) is a long-standing challenge in the vision-language learning domain~\cite{qiao2020referring}. Given an input image and a text description of an object in the image, the task is to localise the object and output its bounding box coordinates. REC challenges models to understand the semantics of the visual and textual input and relate the target referred to in the text to an object present in the image.

\begin{figure}[t!]
    \centering
    \includegraphics[width=\columnwidth,trim={4.7cm 0.5cm 0cm 0.5cm},clip]{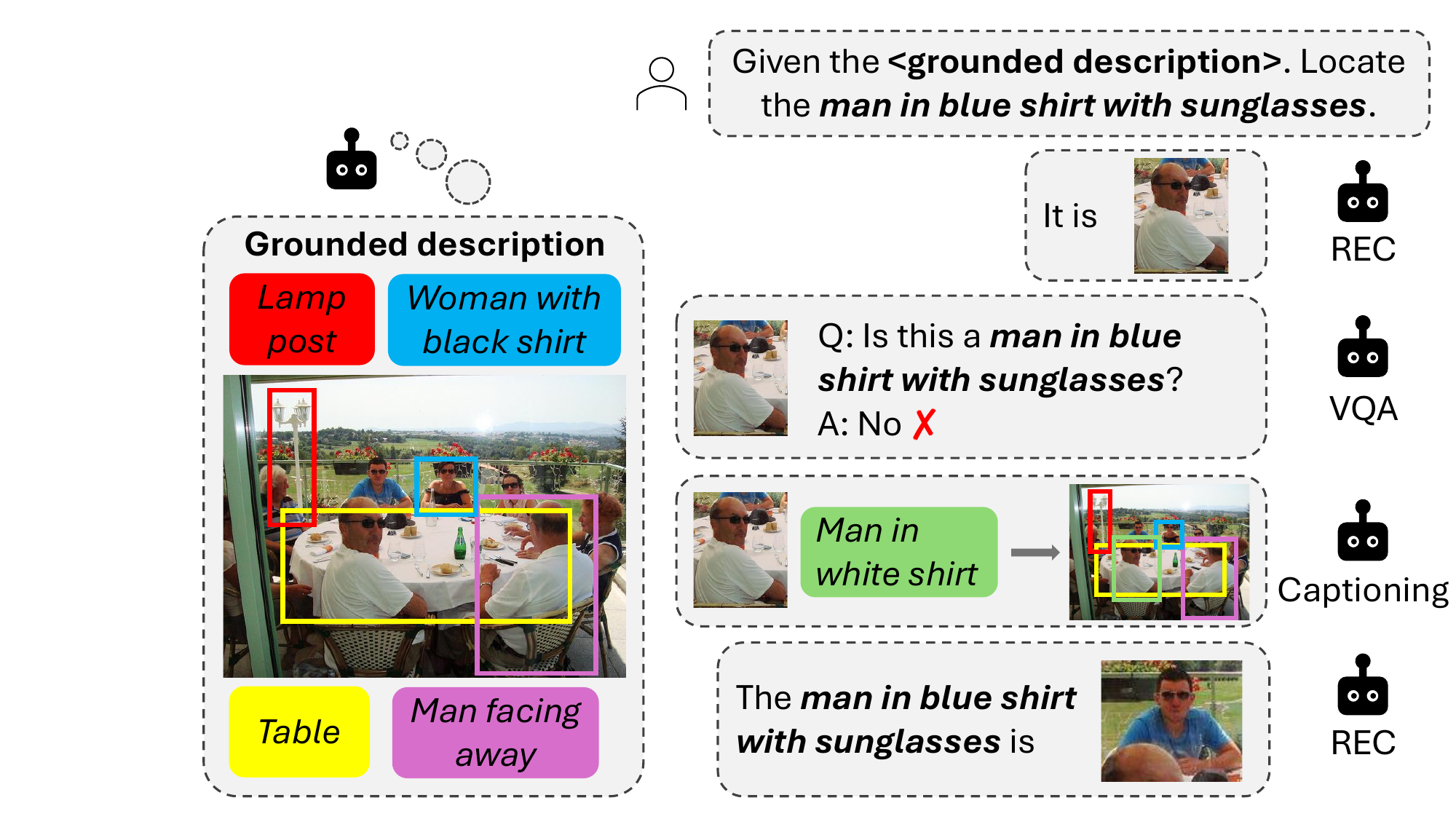}
    \caption{We leverage the multi-task capability of MLLMs, e.g. grounded description generation, referring expression comprehension (REC), visual question answering (VQA), and captioning, to propose a training-free framework for improving REC performance.}
    \label{fig:intro}
    \vspace{-10pt}
\end{figure}

Recent works on multimodal large language models (MLLMs) have achieved over 80\% accuracy at a threshold of 0.5 Intersection over Union (IoU) on various benchmarks for REC~\cite{chen2025revisiting}. However, achieving high accuracy at high IoU thresholds (\textgreater0.7) remains challenging. In the REC benchmarks, the target object is often in the foreground and is the main object in the image. In cases where the target is in the background and partially occluded, the existing models struggle to locate the object. Moreover, the presence of multiple same-class instances of the target object may distract the model from making the correct prediction.

Approaches for the REC task can be separated into specialist and generalist models~\cite{chen2025revisiting}. Specialist models are trained specifically for the REC task, whereas generalist models can perform multiple tasks (including e.g. VQA and image captioning). Although both specialist and generalist models can perform well on the REC task, generalist models, in particular MLLMs, can utilise their multitask capabilities to generate additional context, such as captions and bounding boxes, to further improve their REC performance~\cite{wu2024dettoolchain,fan2025grit}.

In this paper, we dive deeper into the use of tools to provide additional visual and textual context in the task of REC. We conduct experiments on relevant tools for the REC task and investigate their impact on the prediction of generalist models. Furthermore, we aggregate these tools into a training-free framework, named Chain-of-Caption, to improve the performance of existing MLLMs on the REC task. Chain-of-Caption (Fig.~\ref{fig:intro}) focuses on the use of grounded description (i.e. a list of object descriptions and bounding boxes) to relate objects in the image to the textual domain, and iteratively refines the grounded description and target bounding box prediction by leveraging the multi-task capability of MLLMs. We show that our proposed framework can substantially improve the bounding box detection accuracy at a higher IoU threshold (\textgreater0.7) by over 20\%, which results in better aligned bounding boxes\footnote{More visual results are available on the website: \scriptsize\url{https://qm-ipalab.github.io/chain-of-caption/}}. 

\begin{figure*}[t!]
    \centering
    \begin{subfigure}[t]{0.22\textwidth}
        \centering
        \includegraphics[height=1.1in,trim={8cm 5.5cm 16.5cm 6.5cm},clip]{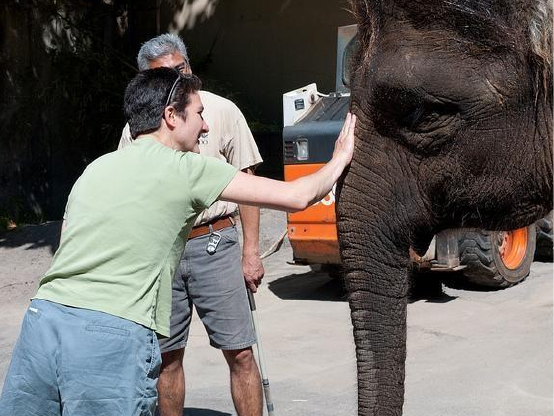}\vspace{-2pt}
        \captionsetup{labelformat=empty}
        \caption{Input image}
    \end{subfigure}
    \begin{subfigure}[t]{0.22\textwidth}
        \centering
        \includegraphics[height=1.1in,trim={0.5cm 0 1.5cm 0.7cm},clip]{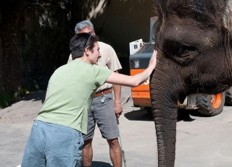}\vspace{-2pt}
        \captionsetup{labelformat=empty}
        \caption{Cropping}
    \end{subfigure}
    \begin{subfigure}[t]{0.22\textwidth}
        \centering
        \includegraphics[height=1.1in,trim={0 0 0 0},clip]{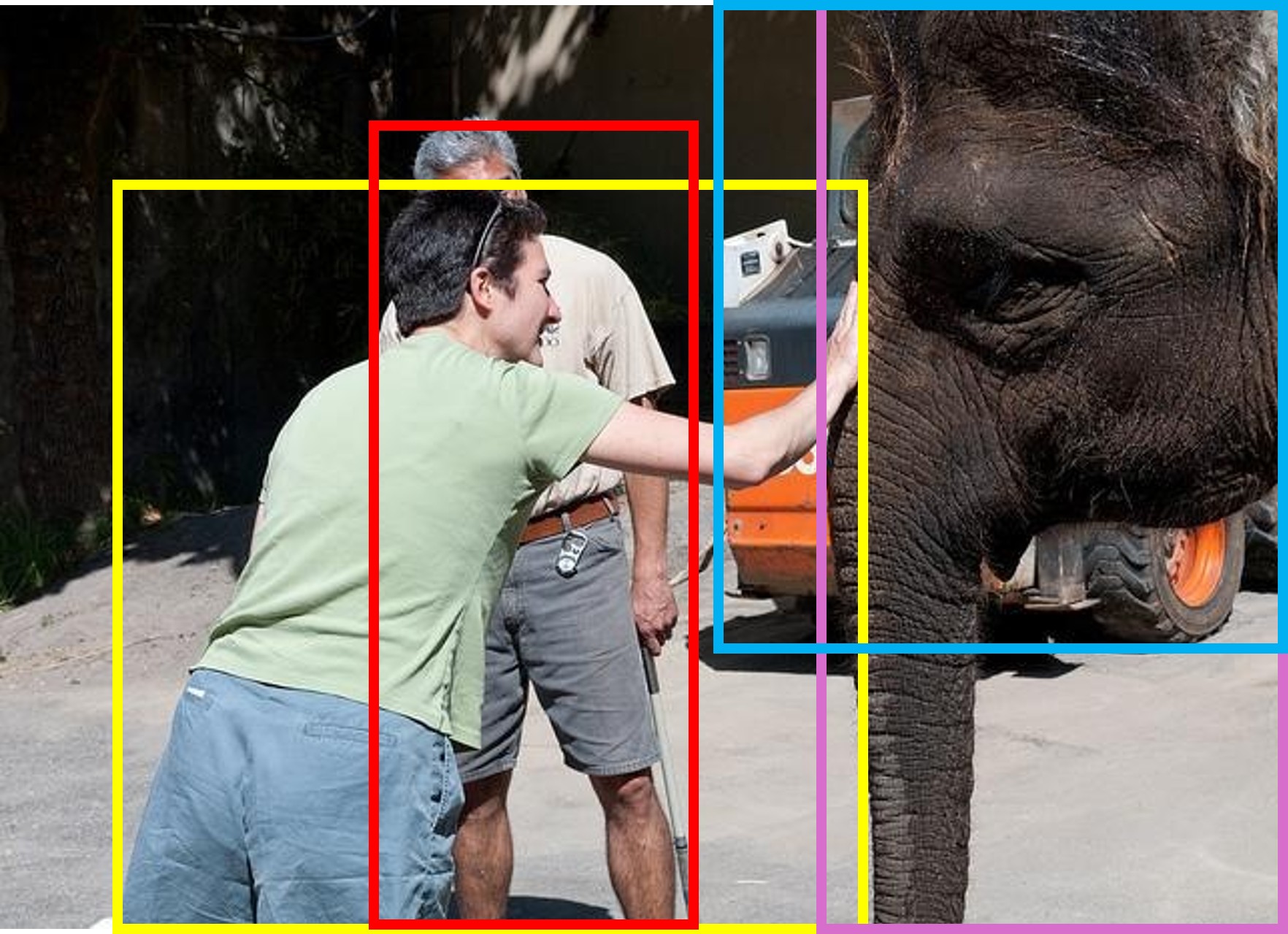}\vspace{-2pt}
        \captionsetup{labelformat=empty}
        \caption{Bounding boxes}
    \end{subfigure}
    \begin{subfigure}[t]{0.27\textwidth}
        \centering
        \includegraphics[height=1.1in,trim={0cm 10.9cm 22cm 0cm},clip]{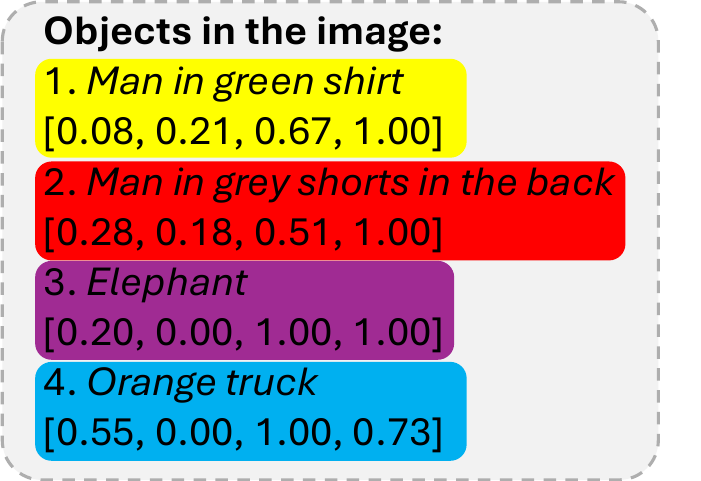}\vspace{-2pt}
        \captionsetup{labelformat=empty}
        \caption{Grounded description}
    \end{subfigure}
    \vspace{-6pt}
    \caption{Example textual and visual contexts for referring expression comprehension. Bounding boxes are in normalised coordinates in the format [top-left x, top-left y, bottom-right x, bottom-right y].}\vspace{-10pt}
\end{figure*}
\section{Related works}
\label{sec:related_works}

Traditionally, models trained for REC are specialist models that can only perform REC and expect unedited images and text descriptions only as input~\cite{mao2016generation,yu2016modeling,luo2017comprehension}. Recent MLLMs are generalist models trained to perform multiple different tasks~\cite{liu2025nvila,bai2025qwen2,liu2024llava}, including REC and other tasks such as image captioning and VQA. MLLMs are often built using auto-regressive modelling strategies which allow them to take unstructured data as input and fuse them together to perform various downstream tasks. Additionally, this auto-regressive capability can be further improved by carefully designed prompts that provide additional information as context for the final prediction~\cite{wei2022chain,shtedritski2023does,yang2023set}.

Since MLLMs can take as input both image and text, context can be provided both in the visual and textual domain by augmenting the input either with external tools or the MLLM itself~\cite{wu2024dettoolchain,fan2025grit,shtedritski2023does,yang2023set}. This is often combined with Chain-of-Thought prompting~\cite{wei2022chain} which allows the model to select which tool to use. DetToolChain~\cite{wu2024dettoolchain} proposes a set of visual tools that can augment the input image to provide guidance to the MLLM on various object detection tasks. The tools include regional amplifiers to allow the model to focus on a specific region of the image, spatial measurement standards to allow the model to measure relative distance in the image, and scene image parsers to highlight potential objects and their relations in the image. The model interactively selects which tool to use at each step to refine the prediction and decide when to end the process. GRIT~\cite{fan2025grit} proposes to use reinforcement learning to train the model to output a reasoning chain that includes language descriptions and object bounding boxes. The bounding boxes help the model relate the text description to the visual information, effectively grounding the reasoning process.

To evaluate REC performance, RefCOCO~\cite{yu2016modeling} and its variants, RefCOCO+/g~\cite{yu2016modeling,mao2016generation}, are often used. The text descriptions in RefCOCO/+ are straightforward, consisting of short descriptions of the target object including appearance and location in the image, while RefCOCOg provides more complicated descriptions in relation to other objects in the scene. However, the RefCOCO/+/g datasets often contain typos and mistakes in the text description. The recently proposed Ref-L4~\cite{chen2025revisiting} dataset provides corrections to the labelling mistakes and extension to the object types from Objects365~\cite{shao2019objects365} and text complexity which provides a better challenge to modern MLLMs.

Although existing approaches show improvements on REC benchmarks, it is not clear where the improvement comes from and why. The different tools are often mixed together in the experiments, making it difficult to understand the contributions of individual components~\cite{wu2024dettoolchain}. GRIT~\cite{fan2025grit} showed that the use of bounding box in the reasoning chain of a fine-tuned model promotes higher attention on the visual tokens overall, indicating that the use of additional context can help the model better relate visual and textual information. In contrast, we study the performance improvements of each individual context in a training-free scenario and propose a combined framework as a result.

\section{Context in referring expression comprehension}
\label{sec:context}

In the task of REC, given an image $x$ and text $t$, an MLLM model $M$ predicts the bounding box $b$ of the target object. By augmenting the text and image in the input, either using preprocessing tools or the MLLM itself, the MLLM can better understand the image and focus on the relevant region for the current text description~\cite{zhang2025mllms,shtedritski2023does,yang2023set}. In this section, we detail techniques for adding visual and textual context that are relevant to the REC task. We subscript the model $M$ to indicate the task that the MLLM is performing (e.g. $M_{\text{REC}}$ is the MLLM performing REC).

\subsection{Textual context}
\label{sec:text_context}
The input text provided in the REC task includes a description of the target object to be detected. This can include the appearance of the object, such as type, colour, shape and size. Although the description can include the rough location of the object in the image (e.g. far left, top right, etc.) and location with respect to other objects within the image (e.g. left of elephant), there is no fine-grained grounding provided (e.g. pixel location or bounding box coordinates).

MLLMs that are trained on grounding tasks can effectively relate bounding box coordinates to locations in the image~\cite{fan2025grit}. Since MLLMs are also capable of describing images in detail including detecting individual objects and generating their description, we prompt the model to detect a specified number of objects in the image and output object descriptions in a numbered list. The model is also prompted to output the bounding box of each object in the same list. This forms a grounded description to allow the MLLM to better relate the text description and the corresponding region in the image. We design a prompt $p$ to generate a grounded description $M_{\text{GDESC}}(x,p)$ as context for the prediction of bounding box $b_{\text{GDESC}}$:

\begin{equation}
    b_{\text{GDESC}} = M_{\text{REC}}(x, M_{\text{GDESC}}(x, p) \oplus t).
\end{equation}

\begin{figure*}[t!]
    \centering
    \includegraphics[width=0.95\textwidth,trim={0cm 5cm 2cm 0cm},clip]{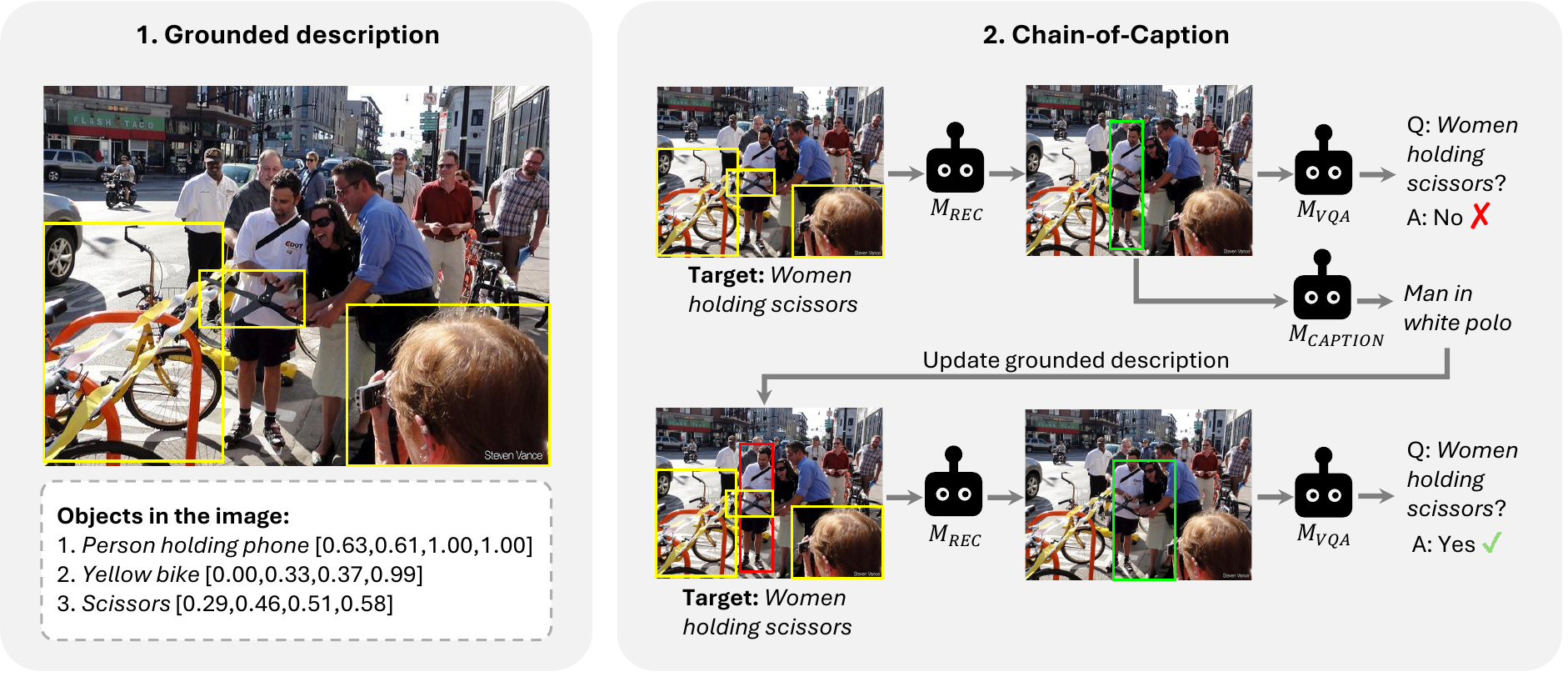}\vspace{-6pt}
    \caption{Our proposed training-free framework for the task of referring expression comprehension. We first initialise the grounded description using the MLLM model. We then refine the predicted bounding box using the multitask capabilities of the MLLM.}
    \label{fig:coc}
    \vspace{-6pt}
\end{figure*}

\subsection{Visual context}
\label{sec:vis_context}
The input image provided in the REC task contains the target object referred to by the text description. The target object can be occluded by other objects making it difficult to detect. Through the use of MLLM, we can augment the image to provide guidance to itself during inference time for the REC task.

Cropping around a region of interest can improve the MLLM's focus on the region~\cite{zhang2025mllms}. We first ask the model to give an initial predicted bounding box $M_{\text{REC}}(x,t)$. We then crop the image to that initial bounding box at 1.5$\times$ the original size and provide the cropped image to the MLLM to obtain the final prediction:
\begin{equation}
    b_{\text{crop}} = M_{\text{REC}}(\text{crop}(x, M_{\text{REC}}(x, t)), t).
\end{equation}
We also draw the bounding box extracted from the grounded description in Sec.~\ref{sec:text_context} as a visual counterpart to the grounded description:
\begin{equation}
    b_{\text{draw}} = M_{\text{REC}}(\text{draw}(x, M_{\text{GDESC}}(x,p)), t).
\end{equation}

\section{Chain-of-caption}
\label{sec:coc}

Adding textual or visual context can guide the model to produce better predictions. Moreover, some contexts are synergistic and combining them together can provide further performance gain. We present Chain-of-Caption, a training free approach to enable the model to ground textual concepts with bounding box coordinates and use cropping and VQA to continuously refine the prediction by appending the caption of failed predictions to the grounded description to yield a more comprehensive context.

We first initialise the grounded description $g=M_{\text{GDESC}}(x, p)$, as described in Sec.~\ref{sec:text_context}. We then combine the input image, grounded description and text, and ask the model to provide an initial predicted bounding box $b = M_{\text{REC}}(x, g, t)$ of the target.

Although the added context of grounded description can improve the bounding box prediction, the prediction can still fail by detecting the wrong object. We rectify the prediction by using the VQA capability of the MLLM to verify whether the predicted region matches the input text description by returning a binary yes/no answer $a$. If $a$ is yes, we terminate the process and output the predicted bounding box $b$. Otherwise, we append the last predicted bounding box with a caption $c$ of the cropped region to the grounded description, and make a new prediction $b_{\text{CoC}}$:

\begin{equation}
    a = M_{\text{VQA}}(\text{crop}(x, b), t),
\end{equation}
\vspace{-6pt}
\begin{equation}
    c = M_{\text{CAPTION}}(\text{crop}(x, b)),
\end{equation}
\vspace{-6pt}
\begin{equation}
    b_{\text{CoC}} = 
    \begin{cases}
        b & \text{if } a=\text{yes}, \\
        M_{\text{REC}}(x, g \oplus (b,c) \oplus t) & \text{ otherwise}.
        
    \end{cases}
\end{equation}

This allows the model to ground the previous predicted region and correct its mistake by making a new prediction with the updated context. We continue the process until the model finds a cropped region matching the input text description (i.e. $a$ is yes) or if the model has made a maximum number of trials. 

\section{Experiments}
\subsection{Datasets}
We use RefCOCO~\cite{yu2016modeling} and its variants RefCOCO+/g~\cite{yu2016modeling,mao2016generation} for our experiments on REC. RefCOCO/+/g consists of images from the MSCOCO dataset~\cite{lin2014microsoft} with text description provided by human annotators. Both RefCOCO and RefCOCO+ datasets consists of short concise text description ($\sim$3.6 words) while RefCOCO contains location information of the object but RefCOCO+ plus does not. Although RefCOCOg contains more complex and longer descriptions ($\sim$8.4 words), it limits the number of objects with the same type to 4 in each image. On the other hand, Ref-L4~\cite{chen2025revisiting} incorporates a cleaned version of RefCOCO/+/g by correcting labelling mistakes, and added instances from the Object365~\cite{shao2019objects365} dataset. The labelling process of Ref-L4 utilises GPT-4V~\cite{gpt4v} to automatically generate captions ($\sim$24.2 words) initially which are then manually reviewed.

\begin{figure}[t]
    \centering
\footnotesize
\begin{tikzpicture}
  \begin{axis}[
    xlabel={Number of objects in grounded description},
    ylabel={Accuracy (\%)},
    xmin=0, xmax=8,
    ymin=0, ymax=70,
    xtick={0,2,...,8},
    ytick={0,20,...,70},
    grid=both,
    width=1.0\columnwidth,
    height=0.5\columnwidth,
  ]

    \addplot[
      color=dc4,
      mark=o,
      mark size=2pt,
      thick
    ]
    coordinates {
      (0,39.76) (1,55.25) (2,57.84) (3,58.17) (4,60.01) (5,61.37) 
      (6,61.28) (7,62.35) (8,62.80)
    };
    \addplot[
        only marks,
        mark=*,
        mark size=2pt,
        color=dc4
    ] coordinates {(0,39.76)};

    \addplot[
      color=dc5,
      mark size=2pt,
      mark=o,
      thick
    ]
    coordinates {
      (0,4.22) (1,19.65) (2,21.62) (3,24.00) (4,26.25) (5,27.17) 
      (6,27.97) (7,27.87) (8,28.5)
    };
    \addplot[
        only marks,
        mark=*,
        mark size=2pt,
        color=dc5
    ] coordinates {(0,4.22)};
    
  \end{axis}
\end{tikzpicture}
\caption{Accuracy improves on RefCOCO when including additional objects in the grounded description as context. Legend: \protect\tikz[baseline]{\protect\draw[dc4, very thick] (0,0.6ex) -- (0.3,0.6ex);} Acc\textsubscript{0.7}, \protect\tikz[baseline]{\protect\draw[dc5, very thick] (0,0.6ex) -- (0.3,0.6ex);} Acc\textsubscript{0.9}, {\protect\raisebox{1pt}{\protect\tikz \protect\draw[black,fill=black] (1,1) circle (0.5ex);}}~w/o grounded description, {\protect\raisebox{1pt}{\protect\tikz \protect\draw[black] (1,1) circle (0.5ex);}}~w/ grounded description.}
    \label{fig:gdesc}
\vspace{-6pt}
\end{figure}
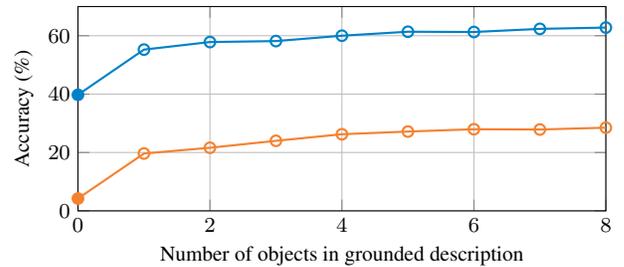

\begin{table*}[t!]
\centering
\footnotesize
\renewcommand{\arraystretch}{1.1}
\setlength\tabcolsep{4pt}

\caption{Comparison of individual contexts and our proposed method Chain-of-Caption.}
\vspace{-6pt}
\begin{tabular}{llcccccccccccc}

\specialrule{1.2pt}{3pt}{0.6pt}
\multirow{2}{*}{Model} & \multirow{2}{*}{Context} & \multicolumn{3}{c}{RefCOCO} & \multicolumn{3}{c}{RefCOCO+} & \multicolumn{3}{c}{RefCOCOg} & \multicolumn{3}{c}{Ref-L4} \\

&&Acc\textsubscript{0.5} & Acc\textsubscript{0.7} & Acc\textsubscript{0.9} & Acc\textsubscript{0.5} & Acc\textsubscript{0.7} & Acc\textsubscript{0.9} & Acc\textsubscript{0.5} & Acc\textsubscript{0.7} & Acc\textsubscript{0.9} & Acc\textsubscript{0.5} & Acc\textsubscript{0.7} & Acc\textsubscript{0.9}  \\

\midrule

\multirow{5}{*}{NVILA-8B} & - & 82.30 & 39.76 & 4.22 & 75.78 & 31.82 & 3.61 & 75.49 & 33.44 & 3.45 & 66.43 & 30.28 & 2.82 \\
& Object description & 79.98 & 41.08 & 4.06 & \textbf{76.97} & 37.98 & 3.84 & 73.11 & 36.79 & 3.28 & 63.95 & 34.35 & 3.20 \\
& Grounded description & 75.88 & 61.37 & 27.17 & 69.98 & 55.85 & 25.97 & 72.25 & 58.46 & 26.52 & 68.42 & 52.80 & 22.44\\
& Cropping & 68.74 & 32.69 & 11.05 & 59.85 & 29.10 & 9.89 & 61.64 & 30.60 & 10.47 & 61.26 & 29.40 & 8.83 \\
& Draw bounding box & 80.48 & 43.56 & 3.42 & 72.91 & 34.45 & 3.36 & 73.46 & 36.23 & 3.02 & 66.02 & 33.73 & 2.96 \\
& Chain-of-caption & \textbf{82.98} & \textbf{68.40} & \textbf{32.50} & 73.59 & \textbf{59.36} & \textbf{29.23} & \textbf{77.45} & \textbf{63.45} & \textbf{30.57} & \textbf{72.11} & \textbf{56.63} & \textbf{25.16}\\

\midrule

Qwen2.5-7B & - & \textbf{87.88} & \textbf{78.24} & \textbf{44.91} & \textbf{80.09} & \textbf{70.20} & \textbf{39.13} & \textbf{83.33} & \textbf{71.93} & \textbf{41.38} & \textbf{81.37} & \textbf{67.04} & \textbf{31.69} \\

\specialrule{1.2pt}{3pt}{1pt}

\multicolumn{2}{c}{\scriptsize{Best scores for each model are in \textbf{bold}.}}
\vspace{-6pt}\vspace{-6pt}
\end{tabular}
\label{tab:ablation}
\end{table*}

\subsection{Discussion}

We first investigate the effect of the number of objects included in the grounded description on the REC performance (Fig.~\ref{fig:gdesc}). The accuracy at higher threshold (0.5 and 0.7) increases with the number of objects included and plateaus as the number of objects reaches above 5. Therefore, we complete the remaining experiments with 5 objects in the grounded description to balance performance and speed.

\begin{figure}[t]
    \centering
    \includegraphics[width=\columnwidth,trim={0cm 9cm 13cm 0cm},clip]{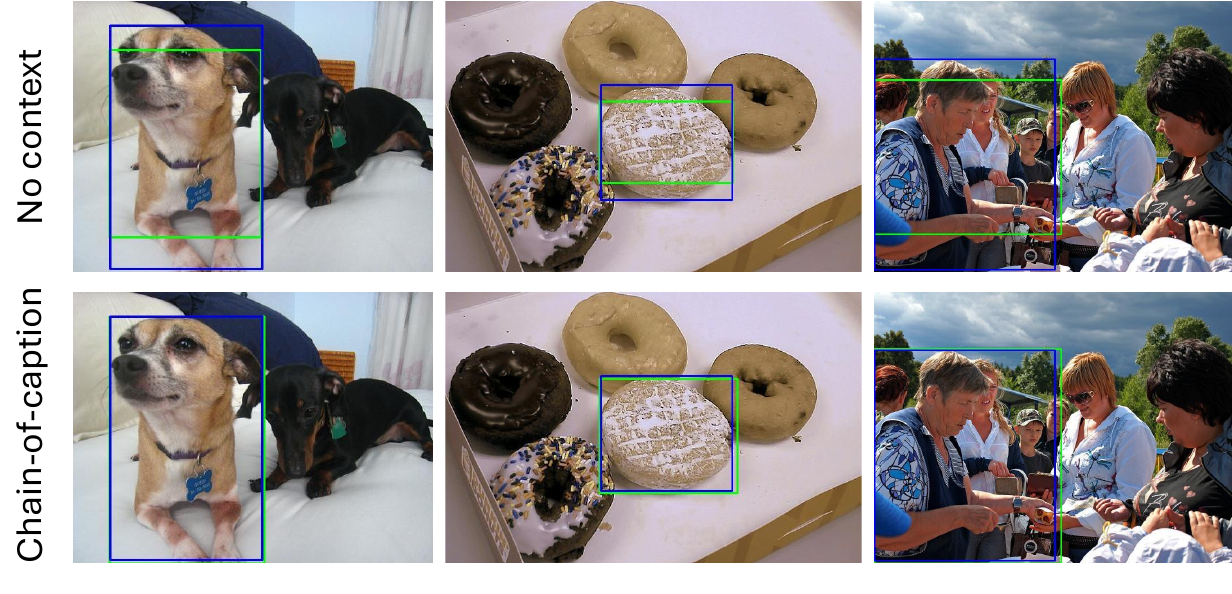}
    \caption{Chain-of-caption refines the predicted bounding box of the base model. \protect\tikz[baseline]{\protect\draw[green, thick] (0,0) rectangle (1.0ex,1.0ex);}~Predicted bounding box, \protect\tikz[baseline]{\protect\draw[blue, thick] (0,0) rectangle (1.0ex,1.0ex);}~Groundtruth bounding box}\vspace{-6pt}
    \label{fig:coc_refine}
    \vspace{-6pt}
\end{figure}

We then study (Tab.~\ref{tab:ablation}) the effect of individual textual and visual contexts using the NVILA-8B model~\cite{liu2025nvila}. We use the accuracy at different IoU thresholds (0.5, 0.7 and 0.9) to evaluate the performance of the model. The base NVILA-8B model exhibits high accuracy (\textgreater80\%) at 0.5 IoU but very low accuracy (\textless5\%) at 0.9 IoU. This suggests that although NVILA-8B is great at predicting where the object roughly is, it is not good at identifying the boundaries of the object, often missing important parts (Fig.~\ref{fig:coc_refine}, top row).

\begin{table}[t!]
\centering
\footnotesize
\renewcommand{\arraystretch}{1.1}
\setlength\tabcolsep{3pt}
\caption{Improvement on the Ref-L4 benchmark by using Chain-of-Caption (CoC) on different model sizes.}
\vspace{-6pt}
\begin{tabular}{lllll}

\specialrule{1.2pt}{3pt}{0.6pt}
Model & Context & Acc\textsubscript{0.5} & Acc\textsubscript{0.7} & Acc\textsubscript{0.9} \\

\midrule

\multirow{2}{*}{NVILA-8B} & - & 66.43 & 30.28 & 2.82 \\
& CoC & 72.11 (\textcolor{green}{+5.68}) & 56.63 (\textcolor{green}{+26.35}) & 25.16 (\textcolor{green}{+22.34})\\

\midrule
\multirow{2}{*}{NVILA-15B} & - & 64.35 & 26.98 & 2.37 \\
& CoC & 78.70 (\textcolor{green}{+14.35}) & 62.90 (\textcolor{green}{+35.92}) & 28.58 (\textcolor{green}{+30.95}) \\

\specialrule{1.2pt}{3pt}{1pt}

\end{tabular}
\label{tab:improve}
\vspace{-6pt}
\end{table}
We explore the effect of textual context using grounded description and also object description which does not contain bounding boxes. Using object description only does not affect the performance much as the accuracy at 0.7 IoU is slightly increased but the accuracy at 0.5 IoU is slightly decreased. However, using grounded description significantly improves the performance at higher IoU of 0.7 and 0.9. This suggests that the grounded description provides effective knowledge of the objects' location in the image and allows the model to focus on the object that best matches the input description. However, one drawback of using grounded description as context is that it decreases the performance at a lower IoU of 0.5.

We also explore the effect of visual context with cropping and visualisation of the bounding boxes obtained from the grounded description. We found that cropping the input image around an initial prediction does improve the performance at a higher IoU of 0.9. This finding aligns with previous works~\cite{zhang2025mllms}, showing that the model is better at locating the object when it is the main focus of the image. However, the performance at a lower IoU of 0.5 is significantly reduced. This may be due to the reduced visual context available after the cropping or an inaccurate initial bounding box. We also studied the effect of visualising the bounding boxes obtained from the grounded description of 5 objects. The results show that there is no significant difference in performance across all datasets.

From the comparison of individual contexts, we conclude that the grounded description is the most effective at enabling the model to ground objects. Moreover, cropping allows the model to focus on the relevant area if the initial prediction is correct. Therefore, we introduce our method, named Chain-of-Caption, which attempts to build upon the initial grounded description by verifying the cropped region of each prediction step. We tested Chain-of-Caption on NVILA-8B~\cite{liu2025nvila}. We found that Chain-of-Caption significantly improves the accuracy at higher IoU of 0.7 and 0.9 over the base model and achieves competitive performance compared to the newer Qwen2.5-7B~\cite{bai2025qwen2}. This can be seen in Fig.~\ref{fig:coc_refine} where we visualise the refined predicted bounding boxes. Moreover, the accuracy at 0.5 IoU is better than the base model in most cases, suggesting that the verification procedure is effective at correcting inaccurate initial bounding box predictions. Furthermore, we tested Chain-of-Caption on different model sizes and Tab.~\ref{tab:improve} shows that our proposed approach is effective at improving the performance of the NVILA-8B and NVILA-15B models. Interestingly, NVILA-15B performed worse than NVILA-8B in its base form, but better after Chain-of-Caption is added, suggesting that NVILA-15B has better in-context learning ability. We also tested Chain-of-Caption on Qwen2.5-7B~\cite{bai2025qwen2}. However, Qwen2.5-7B in its base form is not able to follow the instruction to generate grounded descriptions consistently, often producing outputs with inconsistent format and missing bounding boxes. One way to improve output consistency is to use specific prompts to structure the output in formats such as JSON~\cite{shorten2024structuredrag}.

\label{sec:experiments}

\section{Conclusion}
In this paper, we explored the effect of individual contexts on the performance of MLLMs on REC. We found that the use of grounded description enables the model to effectively relate textual object descriptions to the image, improving the prediction accuracy especially at high IoU thresholds. Furthermore, we proposed Chain-of-Caption, which exploits the grounded description and use the VQA and captioning capability of the MLLM to further improve the contexts and detection performance as a result. Future work includes the use of structured output prompting to improve output consistency.
\label{sec:conclusions}

\bibliographystyle{IEEEbib}
\bibliography{refs}

\end{document}